\documentclass{article}
\usepackage{float}
\usepackage{url}
\usepackage{textcomp}
\usepackage{graphics} 
\usepackage{amsmath}
\numberwithin{equation}{section}
\numberwithin{figure}{section}
\numberwithin{table}{section}
\usepackage{amsthm}
\usepackage{amsmath,amsfonts,amssymb}
\usepackage[linesnumbered,algoruled,boxed,lined]{algorithm2e}
\usepackage{algpseudocode}
\usepackage{mathtools}
\DeclareMathOperator*{\argmax}{arg\,max}
\DeclareMathOperator*{\argmin}{arg\,min}
\newcommand{\dst}{\displaystyle}

\title{Discriminative Dictionary Learning based on Statistical Methods}
\author{{G.Madhuri, Atul Negi}\thanks{School of Computer and Information Sciences, University of Hyderabad, India}}
\date{}
\begin{document}
\maketitle

\begin{abstract}
  Sparse Representation (SR) of signals or data has a well founded theory with rigorous mathematical error bounds and proofs. SR of a signal is given by superposition of very few columns of a matrix called Dictionary, implicitly reducing dimensionality. Training dictionaries such that they represent each class
of signals with minimal loss is called Dictionary Learning (DL). Dictionary learning
methods like Method of Optimal Directions (MOD) and K-SVD have been successfully used in reconstruction based applications in image processing like image ``denoising", ``inpainting" and others. Other dictionary learning algorithms such as Discriminative K-SVD and Label Consistent K-SVD are supervised learning methods
based on K-SVD. In our experience, one of the drawbacks of current methods
is that the classification performance is not impressive on datasets like Telugu
OCR datasets, with large number of classes and high dimensionality. There
is scope for improvement in this direction and many researchers have used statistical methods to design dictionaries for classification. This chapter presents a review of  statistical techniques and their application to   learning discriminative dictionaries. The objective of the methods described here is  to improve classification using sparse representation. In this chapter a hybrid approach is described, where sparse coefficients of input data are generated. We use a  simple three layer Multi Layer Perceptron with back-propagation training as a classifier with those sparse codes as input. The results are quite comparable with other computation intensive methods. 
  
\end{abstract}





\section{Introduction}
Due to immense increase in social media, digital business practices etc., data created, captured, copied or consumed went from 1.2 trillion GB to 59 trillion GB (2010-2020) (Source: Forbes.com, ``54 Predictions About the State of Data in 2021", Gil Press- Forbes). Hence there is a great requirement for faster and efficient methods to categorize or classify data for search or retrieval. In an abstract sense, these methods are well known in literature and are called Pattern classification Methods. Pattern classification involves efficient representation of data as $d-$dimensional feature vectors, designing a discriminant function with classification error as criterion to decide the class membership of a new data vector. Statistical decision theory has been used historically to define the decision boundaries of pattern classes. 
\subsection{Regularization and Dimension Reduction} When the sample size is small compared to the number of variables, any model trained on such data could be overfit i.e. the classification rule learns parameters, noise in the data and hence cannot classify new samples correctly. Regularization is a method to reduce the complexity of a model by decreasing the importance of some variables to zero. 
Retaining relevant features which have variance in the data and dropping features with high correlation or low variance results in reduced dimensionality. Principal Component Analysis (PCA) and Independent Component Analysis (ICA) are applied to attain reduced dimensionality. Non-negative Matrix Factorization (NMF) has been used for dimension reduction in \cite{NMFfordimreduct}.
According to \cite{smallsampleeffectakjain}, to improve the robustness of a classifier in case of few training samples of high dimensionality, features for discrimination and other design parameters such as window size used in Parzen windows approach, number of features used in decision rule, number of neighbours in k-NN method etc, have to be carefully selected. With the rise in online and mobile applications, a mathematically sound model which replaces hand crafted feature extraction and capable of working with limited computational resources and few training samples is of interest. 
\subsection{Sparse representation (SR)} SR has its roots in compressed sensing. Olshausen and Field in \cite{olshausen1997sparseV1?}, proposed that sparse representation model is similar to the receptive field properties of sensory cells in mammalian visual cortex. Field  \cite{field1994goalofsensorycoding} has applied log-Gabor filters on images and the histograms of the resultant output distributions have high kurtosis indicating sparse structure. Field proposed, ``a  high  kurtosis  signifies that  a  large  proportion of the sensory cells is  inactive  (low variance) with  a small proportion of  the cells  describing the contents  of the image (high variance) being active". These works support the idea of sparse representation of natural images. 

 With rigorous proofs and with proven error bounds, sparse representation is a viable model for constrained resource based applications. SR model finds a low dimensional subspace to embed the given high dimensional signals. This embedding is performed against a fixed basis matrix called \textit{Dictionary}. If the dictionary is perfect for the given set of signals, then the input signal or image can be represented with very few columns of the dictionary, with corresponding very few coefficients. 
 
\begin{figure}[h]
     \centering
     \includegraphics[height=9cm,width=12cm]{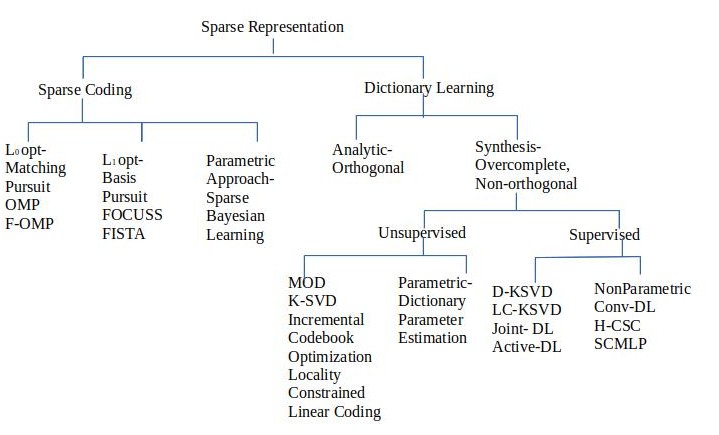}
     \caption{Categorization of Sparse Representation Algorithms}
     \label{treestruc}
 \end{figure}

 Section \ref{notation} describes the notation used throughout the article, Section \ref{spcodingmethods} gives an account of sparse coding methods based on $l_0, l_1$ optimizations and statistical modeling based sparse coding methods. Section \ref{dictlearnmethods} describes the differences between orthogonal, undercomplete and non-orthogonal, overcomplete dictionaries. The similarities and differences between dictionary learning and other subspace learning methods are also discussed in the same section. Section \ref{statsinDL} gives a review of statistical methods used in the design of discriminative dictionaries in a variety of applications like MRI data classification, surgeon classification and level of skill identification based on surgical trial data, histogram feature based supervised dictionary learning for face recognition, etc. Section \ref{nonparaDL} reports usage of CNN based DL for content and style separation in images and generation of new set of images using sparse coding based  Convolutional Neural Network (CNN) and Convolutional Dictionary Learning. Results of using a hybrid dictionary learning method to classify high dimensional data using a simple Multi-Layer Perceptron which is a non-parametric statistical approach, are also discussed here. Section \ref{conclude} concludes. The categorization among various sparse coding algorithms and dictionary learning algorithms is depicted in Fig. \ref{treestruc}.  
 
\section{Notation}\label{notation}
In this section we introduce our notation. 
Throughout the article, $A$ denotes a matrix, $\bf{a}$ denotes a vector, $\|A\|_F$ denotes Frobenious norm of matrix $A$. $A^*$ denotes complex conjugate transpose (Hermitian Conjugate) of $A$ and $A^{\dagger}$ denotes Moore-Penrose pseudo-inverse of $A$.
For a given set of $N$ training patterns, $Y\in {\mathbb{R}}^{d\times N}$ with  dimensionality $d$, the SR model finds a representation of $Y$, 
$Y \approx DX \quad subject to \quad \|\bf{x_i}\|_0 \le T \quad \forall i$, where $X \in {\mathbb{R}}^{K\times N}$ is the \textit{Coefficient Matrix} for input signals.
 $D \in {\mathbb{R}}^{d\times K}$ has $K<<N$ columns $\{\bf d_k\}_{k=1}^K$, called atoms and the matrix is called \textit{Dictionary}. To get non-trivial solutions to this problem, the dictionary atoms are constrained to have  $\|d_k\|_2 \leq 1, \forall k$.  $\|.\|_0$ is the pseudo $l_0-$norm which denotes the number of nonzero components of a vector. $l_p-$norm of a vector $\bf{v}$ is defined as $\dst \|\bf{v}\|_p= \bigg( \sum\limits_{i} |v_i|^p \bigg)^{(1/p)}$. $Y \approx DX$ is
 \begin{equation*}
     \begin{bmatrix}
      \bf{y_1} & \bf{y_2} &\ldots \bf{y_N}
     \end{bmatrix}
     \approx \begin{bmatrix}
      \bf{d_1} & \bf{d_2} &\ldots \bf{d_K}
     \end{bmatrix}
     \begin{bmatrix}
      x_{11} & x_{12} & \ldots x_{1N} \\
      x_{21} & x_{22} & \ldots x_{2N} \\
      \vdots & \vdots & \ldots \vdots \\
       x_{K1} & x_{K2} & \ldots x_{KN} 
     \end{bmatrix}
  \end{equation*} where  $\bf{y_N}$ $\approx$ $x_{1N} \bf{d_1}+ $$ x_{2N}\bf{d_2} + \dots + $$ x_{KN}\bf{d_K}$
 
 Each $x_{kN}$ is the weightage given to dictionary atom $d_k$ in the representation of $N$th training pattern.
 The problem is formulated as an optimization problem in equation \eqref{opteq}
 \begin{equation}\label{opteq}
     Q(Y,D,X) = \argmin \limits_{D,X}\|Y-DX\|_2^2 
 \end{equation}
 subject to $\Arrowvert {\bf{x}_i} {\Arrowvert}_0 \le T, \quad \forall i$. 
 Using Lagrange multiplier method, equation \eqref{opteq} becomes equation \eqref{lagranopteq}.
 \begin{equation}\label{lagranopteq}
     Q(Y,D,X) = \argmin \limits_{D,X}\|Y-DX\|_2^2 + \lambda \|\bf x_i\|_0
 \end{equation}
 The above equation \eqref{lagranopteq} is a nonlinear, non-convex, joint optimization problem which can be solved using Block Coordinate Descent (BCD) method \cite{beck2013convergenceofBCD}. Fixing one variable and updating the other results in two linear optimization problems. Updating the coefficient matrix w.r.t. fixed dictionary is called Sparse coding given by equation (\ref{sparsecoding})
 \begin{equation}\label{sparsecoding}
     Q(Y,D,X) = \argmin \limits_{X}\|Y-DX\|_2^2 + \lambda \|\bf x_i\|_0
 \end{equation}
 This is a combinatorial problem due to pseudo $l_0-$norm, making it a non-convex optimization problem. Convex relaxation of equation \eqref{sparsecoding} is obtained by replacing $l_0-$norm with $l_1-$norm \cite{convexrell0}. 
 \begin{equation}\label{l0l1conrel}
     Q(Y,D,X) = \argmin \limits_{X}\|Y-DX\|_2^2 + \lambda \|\bf x_i\|_1
 \end{equation}
 Equation \eqref{l0l1conrel} is a non-smooth convex optimization problem which can be solved. Dictionary learning problem is discussed in Section \ref{dictlearnmethods}.
 
\section{Sparse Coding Methods}\label{spcodingmethods} 
In this section, we give a broad overview of classification of sparse coding algorithms, based on the norm used for regularization.
Sparse coding algorithms based on $l_0-$norm regularization are easy to implement and thus most popular. Matching Pursuit (MP) \cite{matchingpursuit}, Orthogonal Matching Pursuit (OMP) \cite{OMP}, Fast OMP \cite{fastOMP}, etc., find the sparse coefficient matrix in equation \eqref{sparsecoding}, using a greedy approach. The  coefficient of dictionary atom $\bf{d_k}$ which is highly similar to the input is updated first and the residual after subtracting the contribution of $\bf{d_k}$ multiplied with the coefficient, is again matched with the dictionary atoms. Though these methods work well, they give sub-optimal sparsity levels and sometimes local minima as solutions.

Basis Pursuit (BP) \cite{basispursuit}, Generalised Lasso \cite{lassogeneralized}, Focal Underdetermined System Solver (FOCUSS)\cite{FOCUSS} are some of the important methods of sparse coding using $l_1-$norm optimization in equation \eqref{l0l1conrel}. Such piecewise linear approximations provide a guarantee of maximally sparse unique solution to the sparse coding problem.
A probabilistic model for representing an observed pattern in a lower dimensional space with respect to (w.r.t.) an optimum dictionary and with a prior on the coefficient vector is given in Sparse Bayesian Learning \cite{tipping2001sblrvm}. Sparsity inducing prior acts as a means of regularization. 
Each pattern $\textbf{y}$ is represented as $\textbf{y}=D\textbf{x}+\textbf{n}$ where $\textbf{n}$ is additive Gaussian noise with variance $\sigma^2$. Now, the likelihood function to be maximized is equation \eqref{likelihoodfn}.
\begin{equation}\label{likelihoodfn}
    p(\textbf{y}|D) \propto \int p(\textbf{x})p(\textbf{y}|D,\textbf{x})d\textbf{x}
\end{equation}
Several approximations to equation \eqref{likelihoodfn} have been proposed in \cite{olshausen1997sparseV1?}, \cite{lewicki1999probabilistic}, \cite{lee1999blindsource}, \cite{lewicki2000learningovercomprepresent} to obtain Approximate Maximum Likelihood (AML) estimates of coefficient vector $\textbf{x}$ which maximizes the log likelihood function.
A collection of training patterns 
$\{{\bf y_i}\}_{i=1}^{N}$ are assumed to be independent and different assumptions or approximations about the coefficient vectors 
$\{{\bf x_i}\}_{i=1}^{N}$ result in different estimates. 
For example, in \cite{lewicki1999probabilistic} and \cite{lewicki2000learningovercomprepresent}, components of each coefficient vector are assumed to be independently identically distributed (i.i.d) with a Laplacian prior to promote sparsity i.e. $\dst p(x_i) \propto \exp({-|x_i|/\alpha}) $ where $\alpha$ denotes the parameter diversity. 

\subsection{Importance of Statistical concepts in Sparse Coding Methods}
 Assumptions about data and the sampling method used determine the performance of parametric methods.
Some of the Bayesian sampling techniques used in pattern recognition are Rejection sampling, Ratio of uniforms, Importance sampling, Markov Chain Monte Carlo (MCMC) methods, Slice sampling \cite{bayesamplingmethourl}, \cite{neal2004bayesiansamp}.

In \cite{blumensath2007montecarlosampling}, sparse coefficients of time series data have been estimated using Gibbs sampling and Importance sampling methods. Gibbs sampler cannot explore the entire posterior distribution but takes samples from just a single mode of the posterior distribution. It is difficult for Gibbs sampler to escape local maxima \cite{orthogonalcomponent2010bayesian}. However, Gibbs sampler combined with annealing techniques can help in faster convergence. 

Partially Collapsed Gibbs (PCG) sampler  replaces some conditional distributions with marginal distributions to overcome the limitations of the standard Gibbs sampler as described by Van Dyk and Park in \cite{pcgsampler1} and \cite{pcgsampler2}. 
In \cite{orthogonalcomponent2010bayesian}, Bayesian inference on the unknown parameters corresponding to each sparse coefficient is conducted using samples generated by PCG sampler. These samples asymptotically follow the joint posterior distribution of the unknown model parameters and their hyperparameters. Such samples can closely approximate the joint maximum a posteriori estimate of the coefficients and the dictionary.

Importance sampler is not good for finding sparse approximations as it depends on the proposal distribution used. Importance  sampler samples from a distribution (proposal) and finds the expectations w.r.t. the target distribution. 

\subsubsection{Priors used in Sparse Approximations}
 Generally, the class conditional probability densities (assume features are continuous),  $p(\textbf{y}|C_i)$ are unknown. If the form of the $p(\textbf{y}|C_i)$ is known, but its parameters like mean and variance are unknown, these unknown parameters are estimated if some prior information is known about these parameters and then the Bayes' decision rule is applied. Bayesian framework for estimation of parameters starts with specifying a probabilistic model from which marginal and posterior distributions can be evaluated. When we have large number of training patterns, the general prior applied is Gaussian prior. Though Gaussian prior works very well, sparsity inducing  Laplacian or Cauchy priors act as a way of regularization and allow working with fewer variables than Gaussian case. Jeffrey's prior is invariant w.r.t change of coordinates and hence works well as a prior for scale parameters.
In \cite{bayesianwithpriorsEurasip}, the author has described several priors on the coefficient vector which induce sparsity.
The Generalised Gaussian prior is given by equation \eqref{gengaussianprior}.
\begin{equation}\label{gengaussianprior}
    p(\bf x|\alpha,\beta) = \prod \limits_{j=1}^n \mathcal{GG}{(x_j|\alpha,\beta)}
    \end{equation}
where \begin{equation*}
    \mathcal{GG}{(x_j|\alpha,\beta)} = \frac{\alpha \beta}{2\Gamma(1/\beta)}e^{-\alpha|x_j|^{\beta}}
\end{equation*}
The shape parameter value $\beta = 2$ gives Gaussian prior which corresponds to $l_2-$norm regularization in equation \eqref{opteq}. $\beta=1$ gives Laplacian prior which is equivalent to $l_1-$norm regularization. The scale parameter $\alpha$ squeezes or stretches  and along with location and shape parameters, determines the shape of a distribution. When compared to Gaussian prior, Laplacian prior and those with $0< \beta < 1$ are good sparsity inducing priors. When the application is compression based, a higher level of sparsity is desired.

If the prior and the posterior are from the same family of probability distributions, then the prior is a conjugate prior for the likelihood function. For example, in \cite{orthogonalcomponent2010bayesian}, additive Gaussian noise has variance $\sigma^2$, with Inverse Gaussian prior. Such conjugate priors help in arriving at a closed form posterior, avoiding numerical integration. The authors \cite{orthogonalcomponent2010bayesian} have used PCG sampler to generate samples of the joint probability distribution of model parameters and hyperparameters, where the prior on coefficient vector $\bf x$ is Bernoulli-Gaussian (BG) distribution with parameters $\lambda_k$ and component variance $b_k^2$ for each $x_k$. The hyperprior on $\lambda_k$ is Beta distribution.

A generic method for sparse coding using Bayesian approach is given in Algorithm \ref{spcodingbayesian}. For simplicity, one-dimensional signals $\bf{y}=(y_1,\ldots,y_N)$ are considered and corresponding errors in $\bf{\epsilon} = (\epsilon_1,\ldots,\epsilon_N)$, coefficients of $\bf{y}$, $\bf{x}=(x_1,\ldots,x_K)$ have to be determined using dictionary $D \in \mathbf{R}^{1 \times K}$. In \cite{empiricalpriorwipf2007empirical}, the authors have used a flexible prior based on original data. But, highly sparse priors like Jeffrey's prior result in multimodal posteriors and hence the problem of local optima arises. 

In \cite{cauchypriorspcoding}, the coefficients are assumed to follow Cauchy distribution which is a heavy-tailed distribution and is a member of the Levy-alpha-stable family of distributions. Cauchy proximal operator has been defined and Cauchy Convolutional Sparse Coding algorithm has been proposed to learn sparse coefficients to minimize the representation loss.

For example, Sparse Bayesian Learning (SBL) is a Bayesian approach to find sparse coefficient vectors of given observations. Multiple Snapshot SBL (M-SBL) is used for a dataset of $N$ observations constituting input data $Y$. The corresponding sparse coefficient matrix $X \sim \mathcal{N}{(\bf \mu, \Sigma)}$.

 In \cite{msbl2016DoA}, the authors have assumed Gaussian hyperprior and achieved results comparable to the state-of-the-art. Though Gaussian hyperprior does not induce high level of sparsity, SBL algorithm which achieves maximally sparse solution even with a random dictionary \cite{SBL_Tipping}, is used in DoA estimation. The sparsity level in each coefficient vector is automatically determined at the point of convergence \cite{SBL_visualtracking2005}. 
\begin{algorithm}\caption{Bayesian Sparse Coding Procedure}\label{spcodingbayesian}
\begin{algorithmic}
\Procedure{BayesianSparseCoding}{$\bf{y}$, $D$, \bf{threshold}}
\State $\bf{y}=D\bf{x} + \epsilon$\\
\State {$\dst \textit{posterior} = \frac{\textit{likelihood}\times\textit{prior}}{\textit{evidence}}$

$\dst p(\bf{x}|\bf{y}, \bf{\theta}) \propto p(\bf{y}|\bf{x},\theta_1).p(\bf{x}|\theta_2)$
where $ \dst \bf{\theta}=(\theta_1,\theta_2)$ are hyperparameters.}\\
\State To estimate $\hat{\bf{x}}$, first estimate $\hat{\bf{\theta}}$. Assign a prior to $\bf{\theta}$ with fixed hyper-hyper-parameters, $\bf{\theta_0}$.\\
$\dst p(\bf{x},\bf{\theta}|\bf{y},\bf{\theta}_0
)= \frac{p(\bf{y}|\bf{x},\theta_1)p(\bf{x}|\theta_2)p(\bf{\theta}|\bf{\theta_0})}{p(\bf{y}|\bf{\theta_0})}$\\

\State Full Bayesian Approach: Using \textit{Joint MAP} to estimate $(\bf{x},\bf{\theta})$ 
$(\hat{\bf{x}},\hat{\bf{\theta}})= \argmax_{\bf{x},\bf{\theta}}\{p(\bf{x},\bf{\theta}|\bf{y},\bf{\theta_0})\}$\\
\State OR
\State Using \textit{Evidence Maximization}: Integrate out parameters to estimate hyper parameters $\hat{\bf{\theta}}$ using 
$\dst p(\bf{\theta}|\bf{y},\bf{\theta}_0) = \int p(\bf{x},\bf{\theta}|\bf{y},\bf{\theta}_0)d\bf{x}$\\
Now, $\dst \hat{\bf{\theta}} = \argmax_{\bf{\theta}} \{p(\bf{\theta}|\bf{y},\bf{\theta}_0)\}$ using \textit{MAP or Expectation Maximization (EM) or Maximum Likelihood Estimation}.
\State Bayesian EM: Consider $\bf{y}$ as incomplete data,        
$\bf{x}$ as hidden random variable. $\dst (\bf{y},\bf{x})$ is complete data. $\ln{p(\bf{y},\bf{x}|\bf{\theta})}$ is complete data log-likelihood.
\State E-step: $\dst Q(\bf{\theta},\bf{\theta}^{(k)}) = E_{p(\bf{x}|\bf{y},\bf{\theta}^{(k)})}[\ln{p(\bf{y},\bf{x}|
\bf{\theta})+ \ln{p(\bf{\theta})}}]$\\
\State  M-step: $\dst \bf{\theta}^{(k)} = \argmax \limits_{\bf{\theta}}{Q(\bf{\theta},\bf{\theta}^{(k-1)})}$\\
\State Repeat E-step, M-step until\\
\State $\dst Q(\bf{\theta},\bf{\theta}^{(k)})- Q(\bf{\theta},\bf{\theta}^{(k-1)})\leq threshold $\\
\State $\dst \hat{\bf{\theta}} = \bf{\theta}^{(k)}$
\State Output: $\dst \hat{\bf{x}} = \argmax\limits_{\bf{x}}\{p(\bf{x}|\bf{y}, \hat{\bf{\theta}})\}$ using MAP estimation.
\EndProcedure
\end{algorithmic}
\end{algorithm}

 Another approach to sparse coding is to generate samples from $p(\bf{x}|\bf{y},\bf{\theta})$ and $p(\bf{\theta}|\bf{y},\bf{x})$ for MCMC sampling methods. To estimate $p(\bf{x},\bf{\theta}|\bf{y})$, Gibbs sampler gives $\bf{x} \sim p(\bf{x}|\bf{y},\bf{\theta})$. These samples approximate $\hat{\bf{x}}$. Now, samples $\bf{\theta} \sim p(\bf{\theta}|\bf{y}, \hat{\bf{x}})$, are used to approximate $\hat{\bf{\theta}}$.
 A technical review of Bayesian approaches to sparse coding methods has been given in \cite{spcodingbayesian}.
 
\section{Dictionary Learning Methods.}\label{dictlearnmethods}
The origins of research into dictionary learning are in Independent Component Analysis, ICA.  ICA minimizes the dependence among vector components by imposing independence upto second order \cite{Comon:1994:ICA} i.e., the variables are linear combinations of unknown latent variables which are also assumed to be independent. For a random vector $\bf{v}$ with finite covariance $C_v$, ICA finds a pair of matrices $\{M,N\}$, $N$ being diagonal whose entries are sorted in descending order, such that $C_v=MN^2M^*$.  Similar to dictionary learning, the directions here are also orthogonal, with unit norm constraint on columns of $M$. The entries in dictionary are real numbers and the largest modulus in each column of $M$ is a positive real number. Sparse representation is closely related to ICA with these conditions and hence can be used as a preprocessing tool, just like ICA, before applying Bayesian detection and classification  \cite{Comon:1994:ICA}.

Fixing $X$ obtained from equation \eqref{sparsecoding}, the joint optimization of $D,X$ in equation \eqref{lagranopteq} is reduced to a linear optimization problem using Block Coordinate Descent method \cite{block_coordinatedescent}. Updating dictionary $D$ w.r.t a fixed coefficient matrix $X$ results in equation \eqref{dictlearn} and this learning phase to update $D$ is called Dictionary Learning (DL).
\begin{equation}\label{dictlearn}
     Q(Y,D,X) = \argmin \limits_{D}\|Y-DX\|_2^2 + \lambda \|\bf x_i\|_0
 \end{equation}
where $\|d_k\|\leq 1, \quad k=1,2,\ldots,K$, $K$ is the size of dictionary $D$.

\subsection{Orthogonal Dictionary Learning}
Initially, mathematical transforms were applied on original data columns to get  orthonormal dictionaries called Analytic dictionaries. Such Wavelet dictionaries, Fourier dictionaries have incoherent atoms, orthogonal to each other, hence opted for compression based applications. The level of sparsity achieved is very good with orthogonal dictionaries. Though PCA is capable of capturing major variance in data, minor details which are crucial for discrimination, are not captured. Moreover, the number of significant eigen values is specified by the user. These limitations of PCA could be overcome by a Synthesis dictionary comprising original data as atoms and then iteratively updated such that the representation error is minimal, with better representations and faster convergence \cite{SRsurvey}.

 Though Non-Negative Matrix Factorization (NMF) works well for compressed representations of data, in case of natural images, NMF does not perform well when compared to overcomplete sparse representations \cite{NMF_SR_sleepsignalclassify}.

\subsection{Overcomplete Dictionary Learning}
 Representation of natural images is rich when the redundancy in data is utilised in the form of overcomplete dictionaries. Atoms of an overcomplete dictionary are selected such that their number $K$ is small compared to the data size $N$, but larger than the input dimensionality i.e., $K>>d$. Unlike undercomplete, orthogonal dictionaries, these overcomplete dictionaries are used in reconstruction based applications like image denoising, inpainting where missing or corrupted part of an image is reconstructed.

 Overcomplete dictionary works in contrast with Vector Quantization (VQ), in which each sample is mapped to exactly one prototype. 
 Dictionary learning algorithms could be used to update prototype vectors as in \cite{DLforVQ}, where dictionary learning helps in better quantization of ECG patterns. 
 
\subsection{Structured Dictionary Learning}
When the training set comprises of features along with their class labels, structured dictionaries could be generated. Sub-dictionaries of all classes are grouped together to form a global shared dictionary which represents features shared by all classes. Sub-dictionaries have atoms used to represent features particularly of a specific class. Minimal reconstruction error in equation \eqref{dictlearn} w.r.t sub-dictionaries decides the label of test pattern. When the number of classes increases, computation of structured sub-dictionaries becomes expensive. A single shared dictionary whose atoms have features of each class as well as common features shared by all classes, saves memory and time. In \cite{Yang2011FisherDD}, to learn a discriminative structured dictionary, reconstruction error term is designed such that a given class of data is represented best by the global dictionary and the corresponding class dictionary but not by other class dictionaries. Fisher criterion, i.e., minimal intra-class scatter and maximal inter-class scatter, is imposed on sparse coefficient vectors, making both the coefficients and the dictionary atoms discriminative, leading to better classification results. 
Within-class-scatter is given by $S_w(X) = \sum\limits_{i=1}^C \sum \limits_{x_k\in X_i}(x_k-mean_i)(x_k-mean_i)^T$ and Between-class-scatter is given by $S_B(X)=\sum\limits_{i=1}^C(mean_i-mean)(mean_i-mean)^T,$ where $mean_i$ is the mean sparse coefficient vector of class $i$ and $mean$ is the mean sparse coefficient vector of all the data.  Fisher Discrimination Dictionary Learning (FDDL) uses alternating optimization method with Fisher Discrimination based Sparse Coding in equation \eqref{FDSR}.
\begin{equation}\label{FDSR}
    L(X_i) = \argmin_{X_i}\{r(Y_i,D,X_i)+c_1\|X_i\|_1+c_2g(X_i)\},
\end{equation}
where $g(X_i)=\{tr(S_B(X))-tr(S_w(X_i))+\eta\|X\|_F^2\}$ could be computed by finding the eigen values of the scatter matrices $S_B(X)$ and $S_w(X_i)$. 
Thus, simple statistical concepts used in FDDL, help in learning a discriminative dictionary. Unsupervised and Supervised methods of learning such structured dictionaries are given in Section \ref{unsupDL} and Section \ref{supDL}.

\subsection{Unsupervised Dictionary Learning Algorithms}\label{unsupDL}  Unsupervised algorithms for dictionary learning result in generative or representative dictionaries, usually applied in image denoising, deblurring and inpainting. The missing pixels of an image can be reconstructed with the help of generative dictionaries.   
In each iteration, Method of Optimal Directions (MOD) \cite{MOD_Engan} updates the dictionary by computing pseudo-inverse of coefficient matrix, which causes slow convergence. Another unsupervised algorithm, K-SVD \cite{ksvd} is a generalisation of k-means algorithm, which converges faster due to simultaneous update of both coefficient vectors and dictionary atoms. Only the elements corresponding to non-zero components of coefficient vector are considered to compute residual signal, $E$ and Singular Value Decomposition (SVD) is applied to diagonalize the residual, $E=U \Delta V^T$. The first column of $U$ gives updated atom and the product of first diagonal element and the first row of $V$ gives updated coefficient vector. Retaining only the major part of the signal in the form of few large singular values, effectively reduces noise and gives a better representation of the signals. Incremental Codebook Optimization \cite{mairal2010onlinedictlearn} and Locality Constrained Linear Coding \cite{wang2010locality} are other unsupervised dictionary learning algoritms.

\subsection{Supervised Dictionary Learning Algorithms}\label{supDL} Supervised dictionary learning gives discriminative dictionaries for classification of patterns using labels of patterns in the formulation of objective function. Face recognition in the presence of obstructions and different moods and postures is an important application where discriminative dictionaries are used.
Discriminative-KSVD (DKSVD) learns a discriminative dictionary by incorporating label information into the objective function of K-SVD. 
\begin{equation}\label{dksvd} 
\left\{\begin{array}{l}
<D,W,X>  = \argmin\limits_{D,W,X} \|Y-DX\|_2 + \alpha \|H-WX\|_2 +\beta \|W\|_2,\medskip \\
 \text{sub. to} \quad \|X\|_0 \leq\tau. 
\end{array}\right.
\end{equation}

The matrix $\begin{pmatrix}D \\ \sqrt{\alpha}W\end{pmatrix}$ is always normalized column-wise, so the regularization penalty $\|W\|_2$ can be dropped to get
\begin{equation}\label{finaldksvd}
\left\{\begin{array}{l}
    <D,W,X>  = \argmin\limits_{D,W,X} \bigg \| \begin{pmatrix}Y \\ \sqrt{\alpha}H \end{pmatrix} - \begin{pmatrix}D \\              \sqrt{\alpha}W\end{pmatrix}X \bigg \|_2 ,\medskip\\
    \text{  sub. to } \quad \|X\|_0 \leq \tau.
\end{array}\right.
\end{equation}
The label matrix $H$ is approximated by a classifier matrix $W$ and the coefficient matrix $X$, using alternating optimization (BCD), given by equation \eqref{dksvd}. With $\alpha$, $\beta$ as regularization parameters, K-SVD algorithm is applied to optimize equation \eqref{finaldksvd}.
A similar approach to learning a discriminative dictionary is Label Consistent KSVD (LCKSVD) \cite{LCKSVD}.
If dictionary atoms are coherent, then there is multiple representation problem. So, a compact dictionary is preferred with which similar signals(from  the same  class) can be described by roughly same set of atoms with almost similar coefficients. Application of statistical methods in feature extraction as well as determining the size of dictionary and the dictionary columns, results in better discriminative dictionaries.
\section{Statistical Concepts in Dictionary Learning}\label{statsinDL}
The problem of identifying a dictionary relies on the assumptions of statistical independence and non-Gaussian distribution set as prior \cite{karin_1_l1regn}. 
The ratio of majority and minority class cardinality could be high leading to high misclassification cost. A probablistic model for sparse representation based classification has been given in \cite{classimbalanceprobab}, to address the problem of class imbalance in dataset. A cost sensitive classification rule based on Bayesian framework with sparse coefficients as features has not only improved accuracy but also reduced misclassification cost.

\subsection{Histogram of Oriented Gradients (HoG)}
In case of high dimensionality, feature descriptors are used to avoid unnecessary computations involved in classification. Histogram of Oriented Gradients (HoG) is a feature descriptor used to define an image by the pixel intensities and intensities of gradients of pixels. Gradients define the edges of an image, so extraction of HoG feature descriptor is same as extracting edges. 

Histogram of Oriented Gradients generates gradients at each point of image providing invariance to occlusions, illumination and expression changes. In \cite{hog_grpsp}, group sparse coding with HoG feature descriptors is used to achieve good results on face recognition.

\subsection{Use of Correlation Analysis in Dictionary Learning}
Correlation is the value of association between two independent or one independent and other dependent variables, determined by measuring the Correlation coefficient (Pearson, Kendall, Spearman) and also the direction of their relationship i.e. positive correlation or negative correlation. Quantification of this association involves computing correlation coefficient ranging between $[-1,1]$. In \cite{pearsoncorrcoefftfacerecog}, Pearson product moment correlation coefficient is combined with the sparse reconstruction error of samples for face recognition. While reconstruction error tries to reduce the error between test sample and same class samples, Pearson correlation coefficient maximizes the error between test sample and other class samples, for improved classification results.\\

Canonical Correlation Analysis (CCA) is an extension of bivariate to multivariate analysis. When there are several factors influencing a single outcome, it is multivariate data and the corresponding correlation analysis is called CCA. 
In \cite{structdictlearnbasedoncorr}, the unknown block structure of dictionary is explored using the correlation among dictionary atoms. This method gives control over the size of blocks. Maximum correlation quotient between the test sample and training samples and the reconstruction residual are weighted in the decision function to determine the label of the test signal. \\

\section{Parametric Approaches to Estimation of Dictionary Parameters}
Parametric approaches make some assumptions about the population distribution from which the training data originated. Central Limit theorem is crucial to these assumptions. The theorem states that if sufficiently large number of random samples are drawn (with replacement) from any population with mean $\mu$ and variance $\sigma^2$, then the distribution of sample means will be approximately Gaussian. Whenever there is uncertainty about the probability model of data, Gaussian probability model can be assumed, to derive the population parameters.

Parametric approach to dictionary learning assumes a known distribution from which the columns of dictionary are drawn and tries to estimate the parameters of the distribution, such as size $K$ and the atoms themselves, by using  maximum likelihood maximization to derive mean and covariance of dictionary column distribution. Full posterior estimates are provided using a Bayesian framework, which takes care of uncertainty and unseen data generally observed in biomedical applications.
For representing an observed pattern in a lower dimensional space w.r.t. a coefficient vector and with a prior on the dictionary parameters $\dst \bf{\theta} = (\{d_k\}_{k=1}^K,K)$, the likelihood function to be maximized is equation \eqref{likelihoodfnDL}.
\begin{equation}\label{likelihoodfnDL}
    p(\textbf{y}|\bf{x}) \propto \int p(\textbf{y}|\theta,\textbf{x})p(\theta)d\theta.
\end{equation}

Approximate Maximum Likelihood estimation of an unknown but deterministic dictionary using equation \eqref{likelihoodfnDL} is equivalent to Method of Optimal Directions (MOD) when the noise is assumed to be Gaussian \cite{ILSDLA2007}. In \cite{overcompletedictandsr}, an algorithm to find a joint Maximum A posteriori Probability (MAP) estimate of an unknown random initial dictionary and the corresponding coefficient matrix, is given.

In \cite{dictparaestimate}, a Bayesian Approach has been employed to estimate dictionary atoms and dictionary size $K$ along with the sparse coefficient vector hyperparameters. Additive noise is assumed whose variance is modelled from a gamma distribution with unknown parameters. Each dictionary atom or column has been assumed to be randomly drawn from a uniform distribution with components from $[0,1]$. Such uniform prior is non-informative, so this assumption is equivalent to taking a random initial dictionary, whose columns have unit norm. The coefficient vectors have been modelled as a zero-mean Gaussian where the covariance matrix is determined by hyperparameters which are assumed to be independently gamma distributed.

The dictionary atom parameters, hyperparameters on coefficient vectors, noise variance are determined by approximating to a MAP estimate, obtained by iteratively maximizing the log-posterior density w.r.t. each of them, keeping the others fixed. This approach is equivalent to the Block Coordinate Descent technique employed to optimize equation \eqref{lagranopteq}.\\

A closed form solution to maximizing likelihood function in equation \eqref{likelihoodfnDL} is intractable, but Monte Carlo methods like Gibbs sampler and Metropolis-Hastings are used to approximate closed-form posteriors of dictionary variables \cite{MCMCparadictionaries}. A Markov Chain (MC) is said to be ergodic or irreducible if it is eventually possible to reach every state from each state with positive probability. In  \cite{MCMCparadictionaries}, uniform ergodicity properties of high dimensional Markov Chain which imply convergence to a stationary distribution independent of the initial states, have been discussed.\\

To approximate posteriors of dictionary, Group-wise sampling and aggregation have been used to identify group-wise similar functional brain networks of different persons in \cite{samplingmethodsdictlearn}. Signal sampling and sparse coding on task fMRI data for learning a shared dictionary within a group of persons has  helped in identifying and examining common cortical functional networks at individual level and population level. The authors have used No sampling, random sampling, uniform random sampling, 2-ring and 4-ring sampling methods and the corresponding statistical significance tests have been conducted.\\

Data driven overcomplete dictionaries enable flexible representations of data and the quality of an overcomplete dictionary could be determined using diversity measures like distance between atoms, reconstruction error, coherence among atoms. The Babel measures and entropy from information theory measure the randomness in a system. A high value of entropy denotes spread of atoms in a dictionary \cite{overcompdictentropy}.\\

Active Dictionary Learning updates dictionary atoms from the information in training data, using different strategies. Selecting the most useful sample by uncertainty sampling and by generalization error are classical strategies. The sample whose label cannot be decided is called uncertainty sample and can be decided using posterior probability, margin sampling and entropy based methods \cite{activeDL}.\\

When the samples are complexly structured like trees and sequences, entropy based queries retrieve informative samples for dictionary building.
The uncertainty sample based on entropy is given by 
\begin{equation*}
 y^* = \argmax_{y} (-\sum_i P_{\theta}(labels_i|y)\log P_{\theta}(labels_i|y)),   
\end{equation*} 
where $\theta$ is the set of dictionary parameters.\\

When the training set contains both labeled and unlabeled samples, informativeness of samples could be decided by the probability distribution of class-specific reconstruction error, which determines how well the current dictionary is able to discriminate the sample. \\
In \cite{activediscrimiDL}, the authors have used both reconstruction error of a sample w.r.t. shared dictionary and its entropy on the probability distribution over class-specific reconstruction error, to determine the dictionary. Here, level of discrimination of dictionary is given by the entropy on the probability distribution of error of labeled samples  and level of representation is given by the distribution of the error of unlabeled samples.

\subsection{Hidden Markov Model (HMM)- Discriminative Dictionary Learning}
With Hidden Markov Model, it is possible to describe sparsity profile as each hidden state represents a set of non-zero coefficients. In \cite{hiddenmarkovsparselearning}, the problem of sparse representation has been modeled as a HMM. The approach in this paper has combined filtering based on HMM and manifold-based dictionary learning for estimating both the non-zero coefficients and the dictionary.\\
An equivalence relation, partitioning the set of dictionaries into equivalence classes, has been introduced. A direct search for the equivalence class which contains the true dictionary has been used. The observations are decoupled using a new technique called Change-of-measure, so that the observations are all uniformly, identically distributed. \\

Expectation Maximization has been used to recursively update state in the Markov chain i.e., coefficient matrix $X$ with Gaussian prior, transition matrix of Markov chain and the dictionary.\\

Sparse HMM has been used in \cite{sparseHMMsurgical}, to model surgical gestures, where the dictionary is a set of basic surgical motions. The algorithm to learn a dictionary for all gestures and an HMM grammar describing the transitions among gestures has been proposed here. New motion data is classified based on these dictionaries and grammars. Viterbi algorithm is used for surgeme classification.\\

Given a surgery trial $\{\bf{y_t} \in \mathbf{R}^d\}_{t=1}^T$, assign a surgeme label $v_t \in \{1,2,\ldots,V\}$ to each frame $y_t$. Skill-level from $l \in \{1,2,\ldots,L\}$ is assigned to the trial $\{\bf{y_t}\}_{t=1}^T$ . The surgeme label is a hidden (unobserved)  state modeled as a Markov process with transition probability $q_{v^{`}v} = p(v_t=v|v_{t-1}=v^{`})$. Thus, an observation at time $t$, $y_t$ depends on hidden state $v_t$ through the emission probability density $p(y_t|v_t)$, which is generally assumed to be Gaussian or a mixture of Gaussians.\\

Also, $y_t$ is expressed as a superposition of atoms from a dictionary corresponding to gestures. Hence, $y_t$ depends on another hidden variable $x_t$ i.e., $y_t = D_{v_t}x_t + noise$. \\

For each hidden state $v_t$, a Laplacian prior is imposed on $x_t$, to get a sparse latent variable, given in equation \eqref{laplHMMprior}.\\
\begin{equation}\label{laplHMMprior}
    p(x_t|v_t=v)= (\frac{\lambda_v}{2})^K e^{-\lambda_v\|x_t\|_1},
\end{equation}
where $\lambda>0$ is parameter and $K$ is the size of dictionary $D_{v_t}$ corresponding to $v_t$.
Now, \begin{equation*}
    p(y_t|v_t=v,x_t=x) \sim \mathcal{N}(D_vx,\sigma^2_vI),
\end{equation*}
where $D_v$ is an overcomplete dictionary corresponding to surgeme $v$.

Bayesian Expectation Maximization is applied to learn all the transition probabilities $\{q_{s,s^{`}\}}$ and the parameters of each surgeme model $\Theta_v = (D_v,\sigma^2_v,\lambda_v)$, for each $v \in \{1,2,\ldots,V\}$.\\

To get the surgeme labels $\{v_t\}_{t=1}^T$ of a given trial $\{y_t\}_{t=1}^T$, a dynamic programming approach has been given. If the number of states is finite, then the algorithm converges. \\

For skill-level classification, three Sparse HMM models are learnt for expert, intermediate and novice levels. Level of skill is determined by using Viterbi algorithm \cite{sparseHMMsurgical}.

\section{Non-parametric Approaches to Discriminative DL}\label{nonparaDL}
Unsupervised and Supervised DL algorithms discussed in Section \ref{unsupDL} and Section \ref{supDL} are non-parametric approaches to DL \cite{MCMCparadictionaries}. Parametric dictionaries consider uncertainty in data and avoid local optima. This property of parametric dictionaries improves generalization of sparse representation model. \\
In Supervised Dictionary Learning algorithms  \cite{Mairal_supervisedDL,Yang2011FisherDD,dksvd,LCKSVD}, sparse codes consistent with class labels are generated for both generative and discriminative models. \\
In \cite{Joint_L}, the objective function is formulated combining classification error and the representation error of both labeled and unlabeled data, with a constraint on number of coefficients. All these algorithms are tersely mathematically formulated, tested on datasets for face recognition like Extended YaleB, AR dataset and handwritten numerals data sets MNIST and USPS. \\

If the form of $p(\textbf{y}|C_i)$ is unknown, there are non-parametric approaches like Parzen windows, K-nearest neighbour rule,  Multi-Layer Perceptron (MLP) with back propagation, to estimate $p(\textbf{y}|C_i)$ from the observed data. To improve generalization, data based  methods require huge data. A simple perceptron with one hidden layer is capable of solving any problem (Cybenko's theorem \cite{cybenko1989approximation}). Considered as a non-parametric method to estimate the optimum weights of neural network,  MLP does not make any assumptions about the data and is used to decide boundaries based on the observed data \cite{jain2000statisticalPR}.

With the increase in input dimensionality, the number of hidden neurons increases exponentially. Convolutional neural networks, Deep Belief Networks with several hidden layers are being used in computer vision and pattern recognition, to achieve best classification results. In deep neural networks, where data paucity could affect generalization, auto-encoder is applied for dimensionality reduction. When the training samples are limited and feature extraction is carried out by several hidden layers, there could be problems like vanishing gradient and overfit. The learning time increases as the gradient vanishes in back propagation \cite{hochreiter1998vanishing}.\\
If the feature extraction step of MLP could be replaced with sparse representation, the classifying capability of MLP could be used to classify data with high dimensionality and fewer samples. \\

In \cite{zazo2019convolutionaldictlearn}, a one-to-one correspondence between the sparse coding step, and deep CNNs, has been proposed, representing images using wavelet analysis, sparse coding, and dictionary learning. Dense signal gives the scale, while SR that selects a few dictionary atoms, gives the detail. Hierarchical  convolutional sparse coding (H-CSC)  and Convolutional Dictionary Learning have been used alternatingly, to generate a different set of images combining the content of one set of images with the style of another set of images \cite{seo2020dictionary}. \\

\begin{figure}[H]
  \centering
  \begin{tabular}{@{}c@{}}
    \includegraphics[width=55pt,height=35pt]{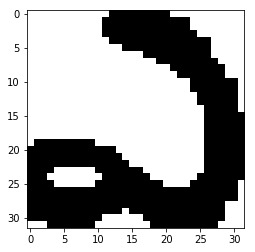} \\[\abovecaptionskip] 
    \small (A)
  \end{tabular}
   \begin{tabular}{@{}c@{}}
    \includegraphics[width=55pt,height=35pt]{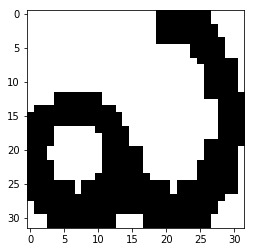} \\[\abovecaptionskip] 
    \small (pa)
  \end{tabular}
  
  \vspace{\floatsep}
  \begin{tabular}{@{}c@{}}
    \includegraphics[width=55pt,height=20pt]{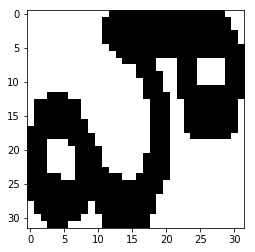} \\[\abovecaptionskip]
    \small (ha) 
  \end{tabular}
\begin{tabular}{@{}c@{}}
    \includegraphics[width=55pt,height=20pt]{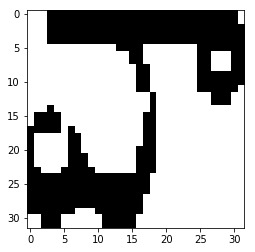} \\[\abovecaptionskip]
    \small (vaa) 
  \end{tabular}
  
 \caption{Similar characters of different classes in UHTelPCC \cite{madhuri2019telugu}.}\label{simichars}
\end{figure}

\begin{figure}[h!]
 \begin{center}
  \includegraphics[height=6cm,width=7cm]{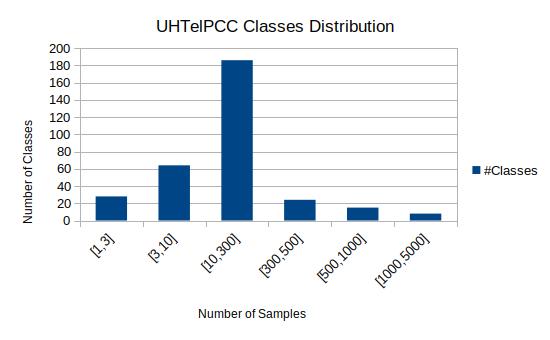}\caption{Number of classes on Y-axis containing number of samples ranging over X-axis \cite{madhuri2019telugu}.}\label{UHbal} 
 \end{center}

 \end{figure}

To overcome the limitations of both Dictionary Learning and Deep Learning, a hybrid method has been proposed, selecting optimal weights and picking the best performing compact architecture empirically, in \cite{madhuri2019telugu}. Sparse coefficients of samples of same class are similar and those of different classes are quite different when computed using a single shared dictionary \cite{LCKSVD}. Here, the authors have used this property of sparse coefficients and Discriminative K-SVD to learn a dictionary to classify datasets which have large number of classes and huge class imbalance ratio. \\
For example, Telugu OCR dataset UHTelPCC \cite{rakeshrtip2r} has high class imbalance as shown in Fig.\ref{UHbal}. Telugu script characters have structural complexity which makes their image feature extraction complex. Also, there is confusing pairs problem as given in Fig.\ref{simichars}, very commonly found in Dravidian scripts.

\begin{figure}[H]
 \begin{center}
  \includegraphics[height=6cm,width=7cm]{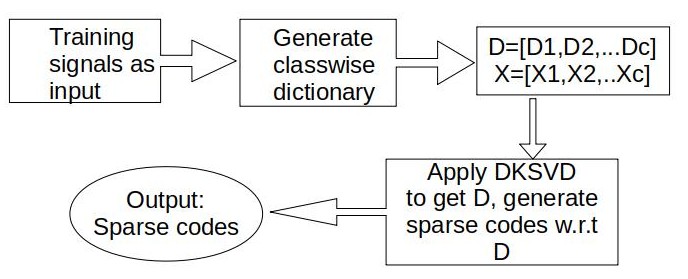}\caption{ Sparse coding w.r.t. shared discriminative dictionary \cite{madhuri2019telugu}.}\label{guconstage1} 
 \end{center}

 \end{figure}

A hybrid method which makes use of the sparse codes as input features avoids tedious feature extraction overhead in deep networks  as shown in Fig.\ref{guconstage1}, leading to a compact MLP architecture. \\

\begin{figure}[H]
 \begin{center}
  \includegraphics[height=6cm,width=7cm]{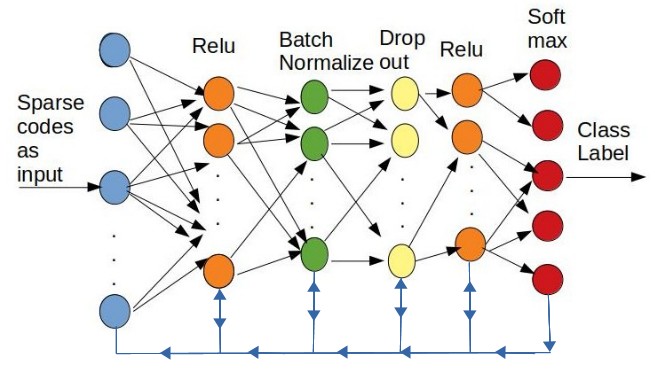}\caption{Training MLP classifier using sparse codes as features \cite{madhuri2019telugu}.}\label{stage2} 
 \end{center}
 \end{figure}

Initialising $W^{(0)} = (X^TX + \lambda I)^{-1}X^TH$, sparse codes generated using equation \eqref{finaldksvd} are given as input to a simple MLP with two hidden layers as shown in Fig.\ref{stage2}.
  The MLP architecture has a dense layer (with ReLU activation), a batch normalization layer, a dropout layer, another dense layer (with ReLU activation). The output layer (with softargmax activation) corresponds to categorical labels of the dataset. 
 The addition of batch normalization layer between hidden layers maps the nonlinear features to the linear part of the activation function.
  Dropout layer has been applied to eliminate the problem of overfit. The MLP is trained on sparse codes generated using DKSVD, and evaluated with sparse codes of test images. Train and test sets of sparse codes are generated w.r.t. same shared dictionary \cite{madhuri2019telugu}.
 \begin{algorithm}\caption{Sparse Code trained MLP (SCMLP)}\label{DLwithDL}
 \SetKwInOut{Input}{input}
 \SetKwInOut{Output}{output}
 \Input{Training signals, $Y_c$,$c=1,2,...,C$, test signals $\textbf x_i$}
 \Output{ Sparse Coefficient matrix $X_{train}, X_{test}$ and validation accuracy}
 For each class of signals $Y_c$, obtain sparse coefficient matrix $X_c$ using a sparse coding algorithm.
 \\After sparse coding, use ApproximateKSVD \cite{Rubinstein_efficientimplementation} to learn class specific dictionary, $D_c, c=1,2,..,C$.
 \\Concatenate class-wise dictionaries and input signals i.e., $D=[D_1,...D_C]$, $Y=[Y_1,Y_2,...,Y_C]$
 \\Apply Discriminative KSVD by adding label matrix H in the objective function and obtain $D$ from equation \eqref{dksvd}.
 \\Extract first $d$ rows of $D$ and normalize to get shared dictionary $D^s$.
 \\For the training set, find the sparse coefficient matrix $X_{train}$ w.r.t. $D^s$ using OMP. Store $X_{train}$.
 \\Feed the sparse codes $X_{train}$ of training signals to the MLP in Fig. \ref{stage2}.
 \\Find the sparse codes $X_{test}$ of test signals w.r.t $D^s$ using OMP and feed them to the trained model to evaluate the performance of the model.
  \end{algorithm}

 Algorithm \ref{DLwithDL} \cite{madhuri2019telugu}, has been tested on UHTelPCC, a printed Telugu connected component dataset and MNIST dataset. 
\subsection{UHTelPCC} 
\begin{figure}[h]
\centering
     \includegraphics[height=6cm,width=7cm]{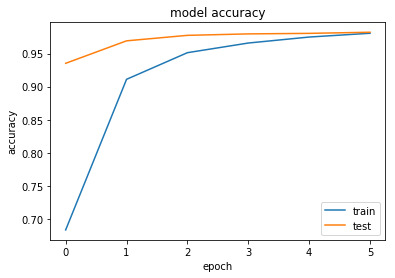}\caption{Model accuracy on UHTelPCC \cite{madhuri2019telugu}.}\label{accuhtel} 
 \end{figure}

\begin{figure}[h]
\centering
  \includegraphics[height=5cm,width=6cm]{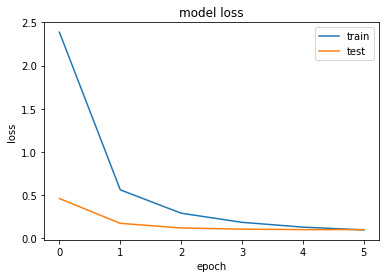}\caption{Model loss on UHTelPCC \cite{madhuri2019telugu}.}\label{lossuhtel} 
\end{figure}

UHTelPCC is a Telugu dataset, contains 70000 binary connected components of size $32 \times 32$ pixels from 325 classes. UHTelPCC is available at\\ http://scis.uohyd.ac.in/~chakcs/UHTelPCC.zip. These 70000 samples are divided into training (50000), validation (10000) and test (10000) sets. Computation times reported in Table \ref{UHTelresults} correspond to training the MLP and validating.  Model accuracy is depicted in Fig. \ref{accuhtel}, Model loss in Fig. \ref{lossuhtel}.

The method has been tested on sparse codes generated from dictionaries of different sizes. The choice of proper size of dictionary for each class is a tradeoff between computing time and accuracy \cite{madhuri2019telugu}. From Table \ref{UHTelresults}, dictionary size of 20 atoms for each class gives 98.7\% accuracy for UHTelPCC with dimensionality 1024. 
\begin{table}[h!]\caption{Results on UHTelPCC dataset \cite{madhuri2019telugu}.}\label{UHTelresults}
\centering 
\begin{tabular}{|c|c|c|c|c|}
 \hline
 \#Atoms & K=16 & K=20 & K=24 & K=26\\
\hline
  Time & 19s & 24s & 27s & 32s\\
\hline
  Accuracy & 97.9 & 98.7 & 98.73 & 98.91\\
\hline
 F1-score & 0.9856 & 0.9963 & 0.9991 & 0.9998\\
\hline
\end{tabular}

\end{table}

\subsection{MNIST} MNIST \cite{ciregan2012multi} is a hand written numerals dataset of 60000 samples for training and 10000 for testing. The dimensionality is 784 and a dictionary size of 18 atoms per class gives 96.3\% accuracy. 

\begin{table}[h!]\caption{Results on MNIST dataset \cite{madhuri2019telugu}.}
\label{MNISTresults}
\centering
\begin{tabular}{|c|c|c|c|c|}
 \hline
 \#Atoms& K=14 & K=16 & K=18 & K=23\\
\hline
  Time & 18s & 21s & 22s & 32s\\
\hline
  Accuracy & 95.34 & 95.12 & 96.32 & 96.4\\
\hline
\end{tabular} 
\end{table}
 Reduced training and testing times for both UHTelPCC and MNIST, from Table \ref{UHTelresults} and Table \ref{MNISTresults}, suggest the low computational complexity of the model. This non-parametric method of learning classifier weights, supports the idea of using statistical concepts in sparse coding as well as dictionary learning.
\section{Conclusion}\label{conclude}
The transformation of dictionary learning from orthogonal transforms to overcomplete analytic transforms to overcomplete synthesis dictionaries is followed by parametric dictionary learning. 
In this review article, we present an overview of using probabilistic models, with different priors and hyper-priors on variables, parametric and non-parametric approaches to parameter estimation, used in sparse representation algorithms. Sampling techniques used in sparse representation to overcome problems like multi-modal data, class imbalance in data, unlabeled data mixed with labeled data and high dimensionality, are discussed. Design of structured, overcomplete dictionaries using entropy analysis of data and examples of research articles presenting Hidden Markov Models for dictionary learning and sparse coding are given. Research articles which combine CNNs with sparse representation to separate content and style in images as well as a hybrid method which combines the representational capabilities of dictionary learning with the classifying capabilities of a neural network are discussed.  

\bibliographystyle{natbib}
\bibliography{bookchap.bib}

\begin{thebibliography}{10}
\expandafter\ifx\csname url\endcsname\relax
  \def\url#1{\texttt{#1}}\fi
\expandafter\ifx\csname urlprefix\endcsname\relax\def\urlprefix{URL }\fi
\expandafter\ifx\csname href\endcsname\relax
  \def\href#1#2{#2} \def\path#1{#1}\fi

\bibitem{NMFfordimreduct}
S.~Tsuge, M.~Shishibori, S.~Kuroiwa, K.~Kita, Dimensionality reduction using
  non-negative matrix factorization for information retrieval, in: 2001 IEEE
  International Conference on Systems, Man and Cybernetics. e-Systems and e-Man
  for Cybernetics in Cyberspace (Cat. No. 01CH37236), Vol.~2, IEEE, 2001, pp.
  960--965.

\bibitem{smallsampleeffectakjain}
S.~J. Raudys, A.~K. Jain, et~al., Small sample size effects in statistical
  pattern recognition: Recommendations for practitioners, IEEE Transactions on
  pattern analysis and machine intelligence 13~(3) (1991) 252--264.

\bibitem{olshausen1997sparseV1?}
B.~A. Olshausen, D.~J. Field, Sparse coding with an overcomplete basis set: A
  strategy employed by v1?, Vision research 37~(23) (1997) 3311--3325.

\bibitem{field1994goalofsensorycoding}
D.~J. Field, What is the goal of sensory coding?, Neural computation 6~(4)
  (1994) 559--601.

\bibitem{beck2013convergenceofBCD}
A.~Beck, L.~Tetruashvili, On the convergence of block coordinate descent type
  methods, SIAM journal on Optimization 23~(4) (2013) 2037--2060.

\bibitem{convexrell0}
S.~Schuler, C.~Ebenbauer, F.~Allg{\"o}wer, l0-system gain and l1-optimal
  control, IFAC Proceedings Volumes 44~(1) (2011) 9230--9235.

\bibitem{matchingpursuit}
S.~Mallat, Z.~Zhang, Matching pursuits with time-frequency dictionaries, IEEE
  Transactions on Signal Processing 41~(12) (1993) 3397--3415.
\newblock \href {https://doi.org/$10.1109/78.258082$}
  {\path{doi:$10.1109/78.258082$}}.

\bibitem{OMP}
Y.~C. Pati, R.~Rezaiifar, P.~S. Krishnaprasad, Orthogonal matching pursuit:
  Recursive function approximation with applications to wavelet decomposition,
  in: Proceedings of the 27 th Annual Asilomar Conference on Signals, Systems,
  and Computers, 1993, pp. 40--44.

\bibitem{fastOMP}
S.-H. Hsieh, C.-S. Lu, S.-C. Pei, Fast omp: Reformulating omp via iteratively
  refining $l<inf>2</inf>$-norm solutions, in: 2012 IEEE Statistical Signal
  Processing Workshop (SSP), 2012, pp. 189--192.
\newblock \href {https://doi.org/10.1109/SSP.2012.6319656}
  {\path{doi:10.1109/SSP.2012.6319656}}.

\bibitem{basispursuit}
S.~Chen, D.~Donoho, Basis pursuit, in: Proceedings of 1994 28th Asilomar
  Conference on Signals, Systems and Computers, Vol.~1, IEEE, 1994, pp. 41--44.

\bibitem{lassogeneralized}
N.~Morioka, S.~Satoh, Generalized lasso based approximation of sparse coding
  for visual recognition, Advances in Neural Information Processing Systems 24
  (2011) 181--189.

\bibitem{FOCUSS}
I.~F. Gorodnitsky, B.~D. Rao, Sparse signal reconstruction from limited data
  using focuss: A re-weighted minimum norm algorithm, IEEE Transactions on
  signal processing 45~(3) (1997) 600--616.

\bibitem{tipping2001sblrvm}
M.~E. Tipping, Sparse bayesian learning and the relevance vector machine,
  Journal of machine learning research 1~(Jun) (2001) 211--244.

\bibitem{lewicki1999probabilistic}
M.~S. Lewicki, B.~A. Olshausen, Probabilistic framework for the adaptation and
  comparison of image codes, JOSA A 16~(7) (1999) 1587--1601.

\bibitem{lee1999blindsource}
T.-W. Lee, M.~S. Lewicki, M.~Girolami, T.~J. Sejnowski, Blind source separation
  of more sources than mixtures using overcomplete representations, IEEE signal
  processing letters 6~(4) (1999) 87--90.

\bibitem{lewicki2000learningovercomprepresent}
M.~Lewicki, T.~Sejnowski, Learning overcomplete representations, Neural
  Computation 12 (2000) 337--365.
\newblock \href {https://doi.org/10.1162/089976600300015826}
  {\path{doi:10.1162/089976600300015826}}.

\bibitem{bayesamplingmethourl}
Sampling methods,
  \url{https://ermongroup.github.io/cs228-notes/inference/sampling/}, accessed:
  03-08-2021 (2021).

\bibitem{neal2004bayesiansamp}
R.~M. Neal, Bayesian methods for machine learning, NIPS tutorial 13 (2004).

\bibitem{blumensath2007montecarlosampling}
T.~Blumensath, M.~E. Davies, Monte carlo methods for adaptive sparse
  approximations of time-series, IEEE Transactions on Signal Processing 55~(9)
  (2007) 4474--4486.

\bibitem{orthogonalcomponent2010bayesian}
N.~Dobigeon, J.-Y. Tourneret, Bayesian orthogonal component analysis for sparse
  representation, IEEE Transactions on Signal Processing 58~(5) (2010)
  2675--2685.

\bibitem{pcgsampler1}
D.~A. Van~Dyk, T.~Park, Partially collapsed gibbs samplers: Theory and methods,
  Journal of the American Statistical Association 103~(482) (2008) 790--796.

\bibitem{pcgsampler2}
T.~Park, D.~A. Van~Dyk, Partially collapsed gibbs samplers: Illustrations and
  applications, Journal of Computational and Graphical Statistics 18~(2) (2009)
  283--305.

\bibitem{bayesianwithpriorsEurasip}
A.~Mohammad-Djafari, Bayesian approach with prior models which enforce sparsity
  in signal and image processing, EURASIP Journal on Advances in Signal
  Processing 2012~(1) (2012) 1--19.

\bibitem{empiricalpriorwipf2007empirical}
D.~P. Wipf, B.~D. Rao, An empirical bayesian strategy for solving the
  simultaneous sparse approximation problem, IEEE Transactions on Signal
  Processing 55~(7) (2007) 3704--3716.

\bibitem{cauchypriorspcoding}
P.~Mayo, O.~Karaku{\c{s}}, R.~Holmes, A.~Achim, Representation learning via
  cauchy convolutional sparse coding, IEEE Access 9 (2021) 100447--100459.

\bibitem{msbl2016DoA}
P.~Gerstoft, C.~F. Mecklenbr{\"a}uker, A.~Xenaki, S.~Nannuru, {Multisnapshot
  sparse Bayesian learning for DOA}, IEEE Signal Processing Letters 23~(10)
  (2016) 1469--1473.

\bibitem{SBL_Tipping}
M.~E. Tipping, \href{https://doi.org/10.1162/15324430152748236}{Sparse bayesian
  learning and the relevance vector machine}, J. Mach. Learn. Res. 1 (2001)
  211–244.
\newblock \href {https://doi.org/10.1162/15324430152748236}
  {\path{doi:10.1162/15324430152748236}}.
\newline\urlprefix\url{https://doi.org/10.1162/15324430152748236}

\bibitem{SBL_visualtracking2005}
O.~Williams, A.~Blake, R.~Cipolla, Sparse bayesian learning for efficient
  visual tracking, IEEE Transactions on Pattern Analysis and Machine
  Intelligence 27~(8) (2005) 1292--1304.

\bibitem{spcodingbayesian}
A.~Mohammad-Djafari, Bayesian approach with prior models which enforce sparsity
  in signal and image processing, EURASIP Journal on Advances in Signal
  Processing 2012~(1) (2012) 1--19.

\bibitem{Comon:1994:ICA}
P.~Comon, \href{http://dx.doi.org/10.1016/0165-1684(94)90029-9}{Independent
  component analysis, a new concept?}, Signal Process. 36~(3) (1994) 287--314.
\newblock \href {https://doi.org/10.1016/0165-1684(94)90029-9}
  {\path{doi:10.1016/0165-1684(94)90029-9}}.
\newline\urlprefix\url{http://dx.doi.org/10.1016/0165-1684(94)90029-9}

\bibitem{block_coordinatedescent}
Y.~Xu, W.~Yin, \href{https://doi.org/10.1137/120887795}{A block coordinate
  descent method for regularized multiconvex optimization with applications to
  nonnegative tensor factorization and completion}, SIAM Journal on Imaging
  Sciences 6~(3) (2013) 1758--1789.
\newblock \href {http://arxiv.org/abs/https://doi.org/10.1137/120887795}
  {\path{arXiv:https://doi.org/10.1137/120887795}}, \href
  {https://doi.org/10.1137/120887795} {\path{doi:10.1137/120887795}}.
\newline\urlprefix\url{https://doi.org/10.1137/120887795}

\bibitem{SRsurvey}
Z.~Zhang, Y.~Xu, J.~Yang, X.~Li, D.~Zhang, {A Survey of Sparse Representation:
  Algorithms and Applications}, IEEE Access 3 (2015) 490--530.
\newblock \href {https://doi.org/10.1109/ACCESS.2015.2430359}
  {\path{doi:10.1109/ACCESS.2015.2430359}}.

\bibitem{NMF_SR_sleepsignalclassify}
M.~{Shokrollahi}, S.~{Krishnan}, Non-negative matrix factorization and sparse
  representation for sleep signal classification, in: 2013 35th Annual
  International Conference of the IEEE Engineering in Medicine and Biology
  Society (EMBC), 2013, pp. 4318--4321.
\newblock \href {https://doi.org/10.1109/EMBC.2013.6610501}
  {\path{doi:10.1109/EMBC.2013.6610501}}.

\bibitem{DLforVQ}
T.~Liu, Y.~Si, D.~Wen, M.~Zang, L.~Lang,
  \href{https://doi.org/10.1016/j.eswa.2016.01.031}{{Dictionary Learning for VQ
  Feature Extraction in ECG Beats Classification}}, Expert Syst. Appl. 53~(C)
  (2016) 129–137.
\newblock \href {https://doi.org/10.1016/j.eswa.2016.01.031}
  {\path{doi:10.1016/j.eswa.2016.01.031}}.
\newline\urlprefix\url{https://doi.org/10.1016/j.eswa.2016.01.031}

\bibitem{Yang2011FisherDD}
M.~Yang, L.~Zhang, X.~Feng, D.~Zhang, {Fisher Discrimination Dictionary
  Learning for Sparse Representation}, 2011 International Conference on
  Computer Vision (2011) 543--550.

\bibitem{MOD_Engan}
K.~Engan, S.~O. Aase, J.~H. Husoy, Method of optimal directions for frame
  design, in: 1999 IEEE International Conference on Acoustics, Speech, and
  Signal Processing. Proceedings. ICASSP99 (Cat. No.99CH36258), Vol.~5, 1999,
  pp. 2443--2446 vol.5.
\newblock \href {https://doi.org/10.1109/ICASSP.1999.760624}
  {\path{doi:10.1109/ICASSP.1999.760624}}.

\bibitem{ksvd}
M.~Aharon, M.~Elad, A.~Bruckstein, {rm K-SVD: An Algorithm for Designing
  Overcomplete Dictionaries for Sparse Representation}, IEEE Transactions on
  Signal Processing 54~(11) (2006) 4311--4322.
\newblock \href {https://doi.org/10.1109/TSP.2006.881199}
  {\path{doi:10.1109/TSP.2006.881199}}.

\bibitem{mairal2010onlinedictlearn}
J.~Mairal, F.~Bach, J.~Ponce, G.~Sapiro, {Online learning for matrix
  factorization and sparse coding.}, Journal of Machine Learning Research
  11~(1) (2010).

\bibitem{wang2010locality}
J.~Wang, J.~Yang, K.~Yu, F.~Lv, T.~Huang, Y.~Gong, Locality-constrained linear
  coding for image classification, in: 2010 IEEE computer society conference on
  computer vision and pattern recognition, IEEE, 2010, pp. 3360--3367.

\bibitem{LCKSVD}
Z.~Jiang, Z.~Lin, L.~S. Davis, Label consistent k-svd: Learning a
  discriminative dictionary for recognition, IEEE Transactions on Pattern
  Analysis and Machine Intelligence 35~(11) (2013) 2651--2664.
\newblock \href {https://doi.org/10.1109/TPAMI.2013.88}
  {\path{doi:10.1109/TPAMI.2013.88}}.

\bibitem{karin_1_l1regn}
R.~Gribonval, K.~Schnass,
  \href{http://dx.doi.org/10.1109/TIT.2010.2048466}{Dictionary identification:
  Sparse matrix-factorization via l1-minimization}, IEEE Trans. Inf. Theor.
  56~(7) (2010) 3523--3539.
\newblock \href {https://doi.org/10.1109/TIT.2010.2048466}
  {\path{doi:10.1109/TIT.2010.2048466}}.
\newline\urlprefix\url{http://dx.doi.org/10.1109/TIT.2010.2048466}

\bibitem{classimbalanceprobab}
Z.~Liu, C.~Gao, H.~Yang, Q.~He, A cost-sensitive sparse representation based
  classification for class-imbalance problem, Scientific Programming 2016
  (2016).

\bibitem{hog_grpsp}
Y.~Li, C.~Qi, Face recognition using hog feature and group sparse coding, in:
  2013 IEEE International Conference on Image Processing, 2013, pp. 3350--3353.
\newblock \href {https://doi.org/10.1109/ICIP.2013.6738690}
  {\path{doi:10.1109/ICIP.2013.6738690}}.

\bibitem{pearsoncorrcoefftfacerecog}
Y.~Xu, J.~Cheng, Face recognition algorithm based on correlation coefficient
  and ensemble-augmented sparsity, IEEE Access 8 (2020) 183972--183982.
\newblock \href {https://doi.org/10.1109/ACCESS.2020.3028905}
  {\path{doi:10.1109/ACCESS.2020.3028905}}.

\bibitem{structdictlearnbasedoncorr}
N.~Kumar, R.~Sinha, Improved structured dictionary learning via correlation and
  class based block formation, IEEE Transactions on Signal Processing 66~(19)
  (2018) 5082--5095.
\newblock \href {https://doi.org/10.1109/TSP.2018.2865442}
  {\path{doi:10.1109/TSP.2018.2865442}}.

\bibitem{ILSDLA2007}
K.~Engan, K.~Skretting, J.~H. Hus{\o}y, Family of iterative ls-based dictionary
  learning algorithms, ils-dla, for sparse signal representation, Digital
  Signal Processing 17~(1) (2007) 32--49.

\bibitem{overcompletedictandsr}
K.~Kreutz-Delgado, B.~Rao, K.~Engan, T.~Lee, T.~Sejnowski, Learning
  overcomplete dictionaries and sparse representations, preparation for
  submission to Neural Computation (2000).

\bibitem{dictparaestimate}
T.~L. Hansen, M.~A. Badiu, B.~H. Fleury, B.~D. Rao, A sparse bayesian learning
  algorithm with dictionary parameter estimation, in: 2014 IEEE 8th Sensor
  Array and Multichannel Signal Processing Workshop (SAM), IEEE, 2014, pp.
  385--388.

\bibitem{MCMCparadictionaries}
T.~Chaspari, A.~Tsiartas, P.~Tsilifis, S.~S. Narayanan, Markov chain monte
  carlo inference of parametric dictionaries for sparse bayesian
  approximations, IEEE Transactions on Signal Processing 64~(12) (2016)
  3077--3092.

\bibitem{samplingmethodsdictlearn}
B.~Ge, X.~Li, X.~Jiang, Y.~Sun, T.~Liu,
  \href{https://www.frontiersin.org/article/10.3389/fninf.2018.00017}{A
  dictionary learning approach for signal sampling in task-based fmri for
  reduction of big data}, Frontiers in Neuroinformatics 12 (2018) 17.
\newblock \href {https://doi.org/10.3389/fninf.2018.00017}
  {\path{doi:10.3389/fninf.2018.00017}}.
\newline\urlprefix\url{https://www.frontiersin.org/article/10.3389/fninf.2018.00017}

\bibitem{overcompdictentropy}
P.~Honeine, Entropy of overcomplete kernel dictionaries, Bulletin of
  Mathematical Sciences and Applications 16 (11 2014).
\newblock \href {https://doi.org/10.18052/www.scipress.com/BMSA.16.1}
  {\path{doi:10.18052/www.scipress.com/BMSA.16.1}}.

\bibitem{activeDL}
J.~Xu, H.~He, H.~Man, Active dictionary learning in sparse representation based
  classification, arXiv preprint arXiv:1409.5763 (2014).

\bibitem{activediscrimiDL}
C.~Zheng, F.~Zhang, H.~Hou, C.~Bi, M.~Zhang, B.~Zhang, Active discriminative
  dictionary learning for weather recognition, Mathematical Problems in
  Engineering 2016 (2016).

\bibitem{hiddenmarkovsparselearning}
L.~Li, A.~Scaglione, Learning hidden markov sparse models, in: 2013 Information
  Theory and Applications Workshop (ITA), IEEE, 2013, pp. 1--10.

\bibitem{sparseHMMsurgical}
L.~Tao, E.~Elhamifar, S.~Khudanpur, G.~D. Hager, R.~Vidal, Sparse hidden markov
  models for surgical gesture classification and skill evaluation, in:
  International conference on information processing in computer-assisted
  interventions, Springer, 2012, pp. 167--177.

\bibitem{Mairal_supervisedDL}
J.~Mairal, F.~Bach, J.~Ponce, G.~Sapiro, A.~Zisserman,
  \href{http://dl.acm.org/citation.cfm?id=2981780.2981909}{{Supervised
  Dictionary Learning}}, in: Proceedings of the 21st International Conference
  on Neural Information Processing Systems, NIPS'08, Curran Associates Inc.,
  USA, 2008, pp. 1033--1040.
\newline\urlprefix\url{http://dl.acm.org/citation.cfm?id=2981780.2981909}

\bibitem{dksvd}
Q.~Zhang, B.~Li, Discriminative k-svd for dictionary learning in face
  recognition, in: 2010 IEEE Computer Society Conference on Computer Vision and
  Pattern Recognition, 2010, pp. 2691--2698.
\newblock \href {https://doi.org/10.1109/CVPR.2010.5539989}
  {\path{doi:10.1109/CVPR.2010.5539989}}.

\bibitem{Joint_L}
D.~Pham, S.~Venkatesh, Joint learning and dictionary construction for pattern
  recognition, in: 2008 IEEE Conference on Computer Vision and Pattern
  Recognition, 2008, pp. 1--8.
\newblock \href {https://doi.org/10.1109/CVPR.2008.4587408}
  {\path{doi:10.1109/CVPR.2008.4587408}}.

\bibitem{cybenko1989approximation}
G.~Cybenko, Approximation by superpositions of a sigmoidal function,
  Mathematics of control, signals and systems 2~(4) (1989) 303--314.

\bibitem{jain2000statisticalPR}
A.~K. Jain, R.~P.~W. Duin, J.~Mao, Statistical pattern recognition: A review,
  IEEE Transactions on pattern analysis and machine intelligence 22~(1) (2000)
  4--37.

\bibitem{hochreiter1998vanishing}
S.~Hochreiter, The vanishing gradient problem during learning recurrent neural
  nets and problem solutions, International Journal of Uncertainty, Fuzziness
  and Knowledge-Based Systems 6~(02) (1998) 107--116.

\bibitem{zazo2019convolutionaldictlearn}
J.~Zazo, B.~Tolooshams, D.~Ba, H.~J.~A. Paulson, Convolutional dictionary
  learning in hierarchical networks, in: 2019 IEEE 8th International Workshop
  on Computational Advances in Multi-Sensor Adaptive Processing (CAMSAP), IEEE,
  2019, pp. 131--135.

\bibitem{seo2020dictionary}
H.-J. Seo, Dictionary learning for image style transfer, Ph.D. thesis, Harvard
  College (2020).

\bibitem{madhuri2019telugu}
G.~Madhuri, M.~N. Kashyap, A.~Negi, {Telugu OCR using Dictionary Learning and
  Multi-Layer Perceptrons}, in: 2019 International Conference on Computing,
  Power and Communication Technologies (GUCON), IEEE, 2019, pp. 904--909.

\bibitem{rakeshrtip2r}
R.~Kummari, C.~Bhagvati, {UHTelPCC: A Dataset for Telugu Printed Character
  Recognition}, in: Recent Trends on Image Processing and Pattern Recognition,
  Vol. 862, 2018, pp. 1--13.

\bibitem{Rubinstein_efficientimplementation}
R.~Rubinstein, M.~Zibulevsky, M.~Elad, Efficient implementation of the k-svd
  algorithm using batch orthogonal matching pursuit, CS Technion 40 (01 2008).

\bibitem{ciregan2012multi}
D.~Ciregan, U.~Meier, J.~Schmidhuber, Multi-column deep neural networks for
  image classification, in: 2012 IEEE conference on computer vision and pattern
  recognition, IEEE, 2012, pp. 3642--3649.

\end{thebibliography}
\end{document}